\documentclass[letterpaper, 10 pt, conference]{latex/ieeeconf}  
\IEEEoverridecommandlockouts                              

\usepackage{graphicx} 
\usepackage[caption=false,font=footnotesize]{subfig}
\usepackage{amsmath}
\usepackage{amssymb}
\usepackage{hyperref}

\usepackage{tikz}
\usepackage{tikzscale}
\usetikzlibrary{calc}

\usepackage{pgfplots}
\pgfplotsset{compat=1.8}

\usepackage{booktabs}
\usepackage{array}

\usepackage{siunitx}

\newcommand{\lvis}{\textsc{Lvis}}

\title{\LARGE \bf
\lvis{}: Learning from Value Function Intervals for Contact-Aware Robot Controllers
}

\author{Robin Deits$^{1}$, Twan Koolen$^{1}$, and Russ Tedrake$^{1}$
\thanks{$^{1}$CSAIL, Massachusetts Institute of Technology, Cambridge, MA, USA
        {\tt\small \{rdeits,tkoolen,russt\}@csail.mit.edu}}%
\thanks{This work was supported by the Fannie and John Hertz Foundation and the MIT Computer Science and Artificial Intelligence Lab. Source code is available at \href{https://github.com/rdeits/LVIS-dev}{https://github.com/rdeits/LVIS-dev}}
}

\begin{document}

\newcolumntype{C}{>{\centering\arraybackslash}m{0.22\textwidth}}

\maketitle
\thispagestyle{empty}
\pagestyle{empty}

\begin{abstract}
Guided policy search is a popular approach for training controllers for high-dimensional systems, but it has a number of pitfalls. Non-convex trajectory optimization has local minima, and non-uniqueness in the optimal policy itself can mean that independently-optimized samples do not describe a coherent policy from which to train. We introduce \lvis{}, which circumvents the issue of local minima through global mixed-integer optimization and the issue of non-uniqueness through learning the optimal value function (or cost-to-go) rather than the optimal policy. To avoid the expense of solving the mixed-integer programs to full global optimality, we instead solve them only partially, extracting intervals containing the true cost-to-go from early termination of the branch-and-bound algorithm. These interval samples are used to weakly supervise the training of a neural net which approximates the true cost-to-go. Online, we use that learned cost-to-go as the terminal cost of a one-step model-predictive controller, which we solve via a small mixed-integer optimization. We demonstrate the \lvis{} approach on a cart-pole system with walls and a planar humanoid robot model and show that it can be applied to a fundamentally hard problem in feedback control---control through contact.
\end{abstract}

\section{INTRODUCTION}
While there are a variety of successful approaches for planning multi-contact behaviors (e.g. \cite{ValenzuelaMixedintegerconvexoptimization2016,DaiWholebodyMotionPlanning2014,Posadirectmethodtrajectory2014,MordatchCombiningbenefitsfunction2014}), it has proven to be difficult to apply these techniques quickly enough to be used in response to disturbances. Furthermore, most multi-contact trajectory optimizations are solved via non-convex optimizations, typically through sequential quadratic programming (e.g. \cite{DaiWholebodyMotionPlanning2014,Posadirectmethodtrajectory2014}) or differential dynamic programming (e.g. \cite{MordatchCombiningbenefitsfunction2014,ZhongValuefunctionapproximation2013,FarshidianRealTimeMotionPlanning2017,LevineGuidedPolicySearch2013}). These techniques can generally find locally optimal solutions, but make no guarantees of global optimality. 
While locally optimal solutions are often sufficient for planning purposes, they make training a policy from examples (as in \cite{LevineGuidedPolicySearch2013} and \cite{ZhongValuefunctionapproximation2013}) more difficult, as the locally optimal samples may not describe a coherent global policy.


Mixed-integer optimization offers some hope for planning globally optimal multi-contact behaviors: By explicitly representing the discrete changes in dynamics with discrete (i.e. integer) variables, we can create optimization problems which are solvable to global optimality using branch-and-bound \cite{FloudasNonlinearmixedinteger1995}. Global optimality is possible even in the presence of nonlinear constraints \cite{BelottiBranchingboundstighteningtechniques2009,Tawarmalanipolyhedralbranchandcutapproach2005}, but for this paper we restrict ourselves to piecewise affine models, inspired by the long history of successful linearized dynamical models for humanoid robots (e.g. \cite{KajitaBipedwalkingpattern2003}). Unfortunately, global optimality comes at a cost, with typical trajectory optimizations taking seconds or minutes to solve \cite{ValenzuelaMixedintegerconvexoptimization2016}. Furthermore, there is no guarantee that these expensive optimizations will result in a consistent global policy, as the optimal policy itself may not be unique.


\begin{figure}[t]
	\centering
	\subfloat[The planar humanoid robot.\label{fig:box-atlas}]{\includegraphics[width=0.23\textwidth]{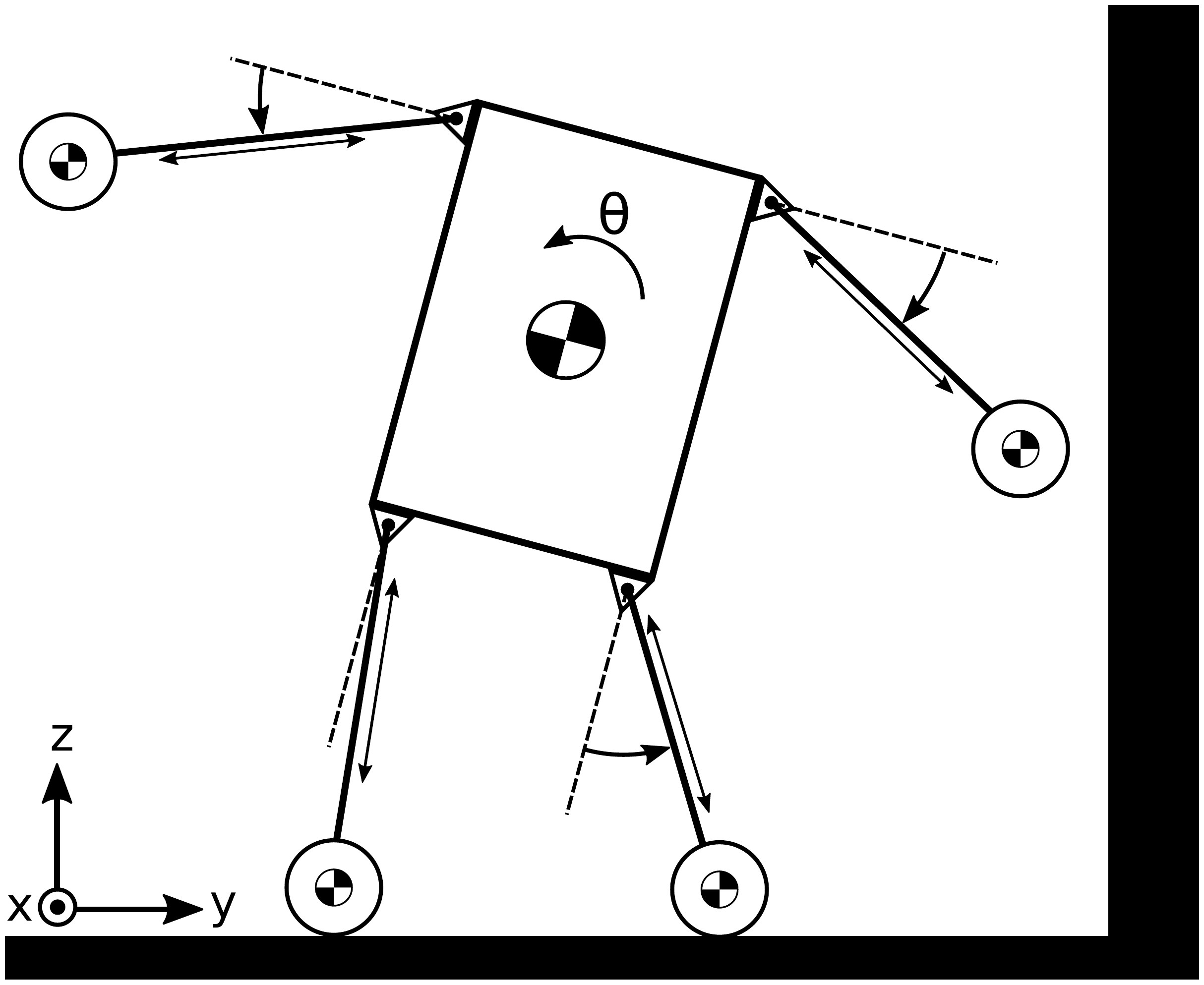}}\quad%
	\subfloat[The cart pole with walls.\label{fig:cart-pole}]{\includegraphics[width=0.23\textwidth]{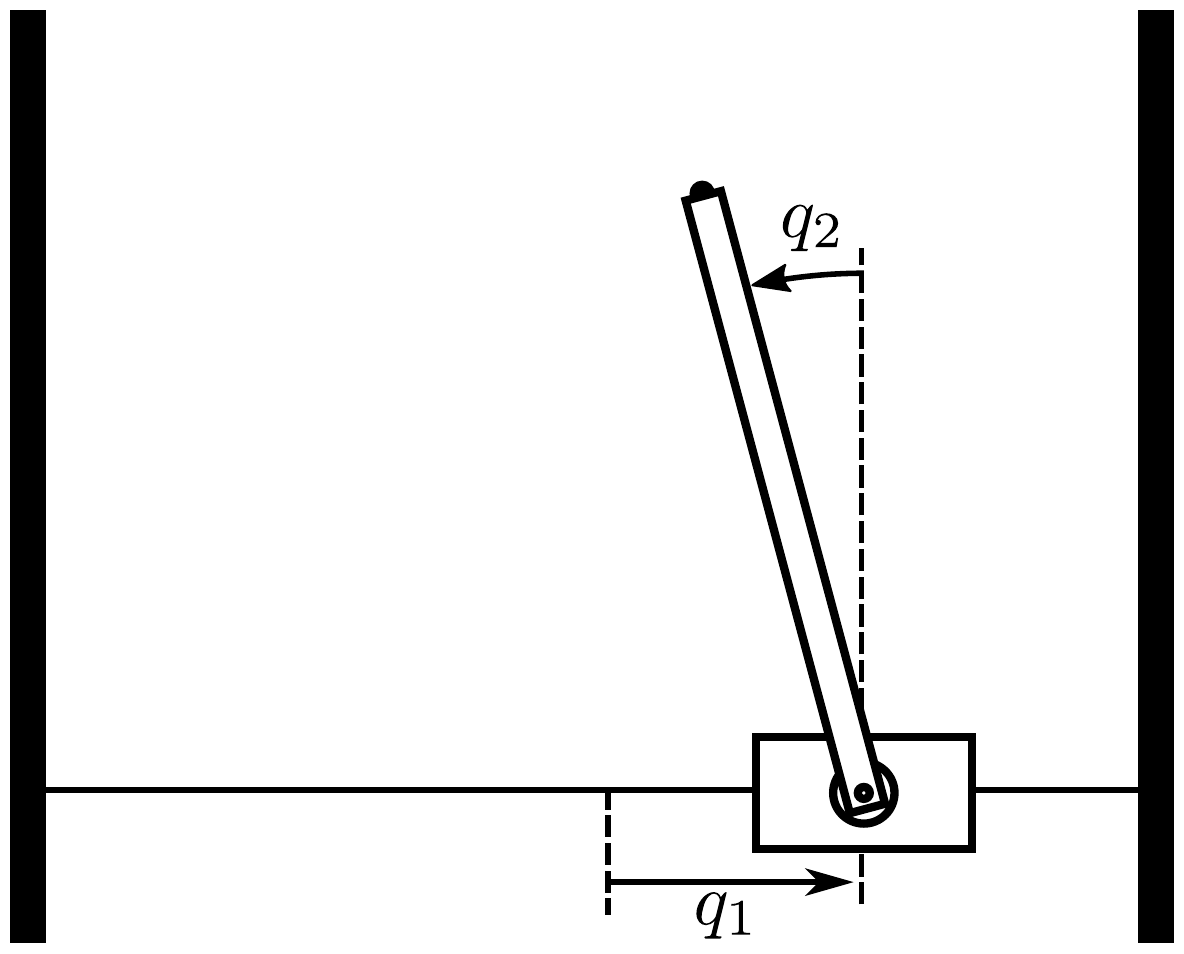}}
	\caption{The robot models used in this work.}
\end{figure}

On the other hand, we do not necessarily need to completely solve a mixed-integer optimization to get some useful information from it. Mixed-integer convex problems are generally solved by branch-and-bound \cite{FloudasNonlinearmixedinteger1995}, a process which iteratively finds better candidate solutions and tighter bounds on the best possible solution. If we ensure that a candidate solution always exists, then we can terminate the branch-and-bound process at any time, retrieving the best solution and tightest bound found so far. Although we could attempt to train from these sub-optimal solutions, we would again be learning to imitate a sub-optimal controller. The \emph{bounds themselves}, however, are extremely useful: In an MPC problem, bounds on the optimal objective value are also bounds on the optimal cost-to-go\footnote{The cost-to-go, which we will also refer to as the value function $J(x)$, is the cost which will be accumulated by the optimal controller starting from state $x$ \cite{TedrakeUnderactuatedRoboticsAlgorithms2018}.} from a given state.\footnote{This assumes that the MPC horizon is long enough to reach a set of states with known cost-to-go. We will violate that assumption later, but attempt to demonstrate empirically that the objective bounds are still useful.} Having a model of the cost-to-go in turn enables fast online control by simply greedily descending that cost. 

\subsection{\lvis{}: Learning from Value Interval Sampling}

In this work, we introduce \lvis{}, a new approach for the creation of contact-aware controllers. We model our robot's contact dynamics with complementarity constraints (Sect.~\ref{sec:complementarity-model}). Offline, we set up a large number of trajectory optimizations in the form of mixed-integer quadratic programs (MIQPs) from a variety of initial states. We partially solve those optimizations, terminating early and extracting concrete intervals containing the optimal cost at the given robot state (Sect.~\ref{sec:learning-mpc-data-collection}). From these intervals, we train a small neural net to approximate the cost-to-go using a loss function which penalizes predicted values outside the known intervals (Sect.~\ref{sec:training}). Online, we run a simple one-step MPC controller to greedily descend the approximate cost-to-go as quickly as possible subject to the robot's dynamics (Sect.~\ref{sec:online}).

\section{RELATED WORK}

This work is similar to that of Zhong et al. in \cite{ZhongValuefunctionapproximation2013}, in which offline optimizations were also used to train an approximate cost-to-go used as the terminal cost of a shorter-horizon MPC problem. Zhong's work differs from ours in its use of iterative LQR (iLQR) to generate the cost-to-go samples. As iLQR is a local nonlinear optimization, it can only provide an estimate of the upper bound of the the cost-to-go (since a lower cost might exist in a space that was not explored by the local optimization). In our case, by constructing a mixed-integer optimization and solving it with branch-and-bound, we recover global upper and lower bounds on cost-to-go, using the interval spanned by those bounds during training. In Sect.~\ref{sec:interval-importance}, we specifically compare \lvis{} with an approach of learning only from upper bounds on the cost-to-go.


A major obstacle to solving MPC problems for system with contacts is the potentially vast number of possible mode sequences.\footnote{Our humanoid system has $1,679,616$ modes (Sect.~\ref{sec:humanoid-model}), so a trajectory optimization with a horizon of 10 steps has $1679616^{10} \approx 1.8\times10^{62}$ possible mode sequences, most of which are infeasible.} If an optimal mode sequence for a given state could be computed, then we could perform a cheap continuous optimization to choose the precise optimal input given that mode sequence. This is the approach taken by Hogan in \cite{HoganReactivePlanarManipulation2018}, in which a neural net is trained to predict mode sequences from robot states. Marcucci takes a similar approach in \cite{MarcucciApproximatehybridmodel2017} by creating a library of provably feasible stabilizing mode sequences and looking up a mode sequence for the robot's current state at run-time.

Alternatively, the efficient sequential linear-quadratic methods from \cite{FarshidianRealTimeMotionPlanning2017} do allow for locally optimal real-time MPC for systems with contact, avoiding the need for offline learning. These optimizations are still subject to local minima, but the ability to run them at real-time means that they do not need to be used to train a global policy.


Looking more broadly, reinforcement learning offers an alternative approach which does not require any explicit offline planning but instead simply the ability to roll out actions in simulation or hardware (e.g. \cite{KoberReinforcementLearningRobotics2013,GuDeepReinforcementLearning2016}). 
In our approach, directly measuring the cost-to-go intervals from a given state, rather than trying to estimate a reward based on expected future actions, allows us to use a very simple supervised learning architecture instead.

\section{Robot Models}

We demonstrate the \lvis{} controller on two models: a cart-pole system balancing between two walls and a simplified humanoid model.
The cart-pole (Figure~\ref{fig:cart-pole}) consists of an actuated cart which can accelerate in one dimension, and an unactuated pole which rotates freely. We modify the system by adding two walls on either side (contact is considered only between the tip of the pole and the walls). The cart-pole system has 4 continuous states, 1 input, and 7 discrete modes (contact-free and sticking, sliding up, and sliding down for each wall). 

\label{sec:humanoid-model}

The simplified planar humanoid model (Figure~\ref{fig:box-atlas}) has 11 degrees of freedom (3 DoF for the planar translation and rotation of the central body, and 2 DoF for the rotation and extension of each limb in the plane). We model contact between each limb and the fixed floor or wall, and we add hard position limits for each of the 8 joints connecting the limbs to the body. The humanoid has 22 continuous states and 11 inputs. Since each of the 4 limbs has 3 states (free, sticking, or sliding in one of 2 directions), and each of the 8 joints has 3 states (free, at its upper limit, or at its lower limit), the system has a total of $4^4 \times 3^8 = 1,679,616$ discrete modes. We model these modes implicitly using complementarity conditions (Sect.~\ref{sec:complementarity-model}). 

\section{TECHNICAL APPROACH}


\subsection{Modeling}
\label{sec:complementarity-model}

Following the formulation of Stewart and Trinkle in \cite{Stewartimplicittimesteppingscheme2000}, we model the dynamics of our robot in a contact-implicit manner with complementarity conditions. In discrete time, these dynamics take the form:
\begin{subequations}
\begin{align}
\mathbf{M} \left(\mathbf{v}^{l+1} - \mathbf{v}^{l}\right) &= h \mathbf{f}_{ext}^l + h \mathbf{C} + h \mathbf{B} \mathbf{u}^l \label{eq:dynamics-v} \\
\mathbf{q}^{l + 1} - \mathbf{q}^{l} &= h\mathbf{v}^{l + 1} \label{eq:dynamics-q}
\end{align}
\end{subequations}
where $h$ is the time step, $\mathbf{q}^l$ and $\mathbf{v}^{l}$ are the system's generalized configuration and velocity at time step $l$, $\mathbf{M}$ is the mass matrix, $h \mathbf{f}_{ext}$ is the external impulse due to contact and friction, $h\mathbf{C}$ is the impulse caused by Coriolis, and gravitational forces, and $h\mathbf{B}\mathbf{u}^l$ is the impulse caused by the generalized inputs $\mathbf{u}$. 
Complementarity conditions ensure that there is no force acting at a distance and no sliding frictional force without an accompanying velocity. For example, we constrain the normal force of a contact with a condition of the form:
\begin{align}
f_{\perp} \perp \phi(\mathbf{q})
\label{eq:simple-complementarity}
\end{align}
where $f_{\perp}$ is the normal force and $\phi(\mathbf{q})$ is the separation between the associated contact point and the world. We use the notation $\mathbf{a} \perp \mathbf{b}$ to mean $\mathbf{a} \geq 0$ and $\mathbf{b} \geq 0$ and $\mathbf{a}^\top \mathbf{b} = 0$, so condition (\ref{eq:simple-complementarity}) has the effect of ensuring that normal force is nonnegative (no pulling on the ground), separation is nonnegative (no penetration), and contact force can only occur when separation is zero. For the complete set of complementarity conditions used in this work, see \cite{Stewartimplicittimesteppingscheme2000}.


It is important to note that $\mathbf{M}$, $\mathbf{C}$, $\mathbf{B}$, and $\phi$ all depend on the robot's current state (in general in a nonlinear way). When we write down a trajectory optimization as an MIQP, we cannot represent these nonlinear dependencies, so we currently linearize the dynamics around the current state. This limits our current implementation, as our results are only valid for the particular linearization. While linearized centroidal dynamics have been dramatically successful in humanoid robotics (e.g. \cite{KajitaBipedwalkingpattern2003}), we hope to explore using the full nonlinear dynamics in future work (see Sect.~\ref{sec:future-minlp}). 

\subsection{Data Collection via Optimal Control}
\label{sec:learning-mpc-data-collection}

To generate a single sample, we start from some initial state $\mathbf{x}^0 = \begin{bmatrix}\mathbf{q}^{0\top} & \mathbf{v}^{0\top}\end{bmatrix}^\top$. Given a convex quadratic cost described by matrices $\mathbf{Q}$, $\mathbf{R}$, and $\mathbf{S}$, we write down a trajectory optimization:
\begin{align}
\begin{split}
J^* = \underset{%
\substack{\mathbf{x}^1 \dots \mathbf{x}^N, \mathbf{u}^1 \dots \mathbf{u}^N, \\ \mathbf{f}^1 \dots \mathbf{f}^N}}%
{\text{minimize}} & \sum_{l=1}^N \left[ \mathbf{x}^{l \top} \mathbf{Q} \mathbf{x}^l + \mathbf{u}^{l \top} \mathbf{R} \mathbf{u}^l \right] + \mathbf{x}^{N \top} \mathbf{S} \mathbf{x}^N\\
\text{subject to} \quad & \text{linearized dynamics (\ref{eq:dynamics-v}, \ref{eq:dynamics-q})}\\
& \text{complementarity (\ref{eq:simple-complementarity})}\, .
\label{eq:miqp}
\end{split}
\end{align}
For each scalar complementarity condition $x_i \perp y_i$ we introduce a new binary variable $z_i$ and constrain:
\begin{subequations}
\begin{align}
x_i &\geq 0, \quad y_i \geq 0 \\
z_i = 1 &\implies x_i = 0  \label{eq:implication-1} \\
z_i = 0 &\implies y_i = 0 \label{eq:implication-0} \, .
\end{align}
\end{subequations}
We formulate the implication constraints in (\ref{eq:implication-1}, \ref{eq:implication-0}) as linear constraints using a standard big-M formulation \cite{LofbergBigMConvexHulls2012}. The introduction of the binary variables $z_i$ converts our optimization into an MIQP, i.e. a program with a quadratic objective, linear constraints, and some variables constrained to take integer values in $\left\{0, 1\right\}$. 

\subsubsection{Generating Feasible Solutions as Warm-Starts}
\label{sec:warm-start}
Because the only constraints present in the optimization problem in (\ref{eq:miqp}) are those that encode the dynamics and complementarity conditions, any desired behavior of the system must be expressed in the cost matrices $\mathbf{Q}$, $\mathbf{R}$, and $\mathbf{S}$. This restricts the expressiveness of our optimization, but it also means we can generate \emph{feasible} solutions to the optimization (\ref{eq:miqp}) simply by \emph{simulating} the system under any controller subject to the same physical constraints. These simulated trajectories are used as warm-starts in the mixed-integer optimization. 

The quality of the feasible solutions we generate depends entirely on the choice of controller used during the simulation. Fortunately, since simulation is computationally cheap compared to a full MIQP solution, we can afford to warm-start with multiple controllers and pick whichever simulation result happens to have lower cost. In our case, we warm-start every optimization by simulating with both a basic LQR controller and the \lvis{} controller being trained. As we train the \lvis{} cost-to-go, its performance improves and it tends to provide increasingly good warm-starts. 

\subsubsection{Early Termination}
\label{sec:early-termination}

Solving the MIQP optimization (\ref{eq:miqp}) can be extremely expensive for a robot with as many states and modes as our planar humanoid: a trajectory with just $N=10$ steps can take thousands of seconds to solve to near optimality\footnote{For our humanoid robot model, solving to within 1\% of the optimal cost required an average of 1160 seconds per trajectory optimization, with some cases taking several hours each. At a nominal control rate of 100\,Hz, solving to full optimality would thus require the data collection to run at less than 0.0008\% real-time.}. The key insight of \lvis{} is that we do not need to solve all the way to optimality: We can easily generate feasible solutions by simulation, so the process of solving the optimization problem in (\ref{eq:miqp}) is a matter of running the branch-and-bound algorithm to iteratively find better solutions and tighter bounds on the optimal cost. At any point, we can simply terminate the optimization and extract the best solution and tightest bound found so far. We label the cost of the best solution found so far (which is an upper bound on $J^*$ as $J_{ub}$, and we label the tightest lower bound on the optimal cost as $J_{lb}$.

We somewhat arbitrarily choose to terminate the optimization at a fixed time limit of 3 seconds to balance the sub-optimality of each sample with the rate at which samples can be generated (limits of 5 or 10 seconds provided similar performance). Other time limits or termination strategies are certainly possible but have not yet been explored. 


\subsection{Training the Neural Net}
\label{sec:training}

The neural net which will approximate our cost-to-go consists of a simple fully-connected feed-forward network with exponential linear unit (ELU) activations \cite{ClevertFastAccurateDeep2015}. For the humanoid in Fig.~\ref{fig:box-atlas} we use two hidden layers with 48 units each, while for the cart-pole in Fig.~\ref{fig:cart-pole} we use two hidden layers with 24 units each. The neural net has a number of input dimensions equal to the number of states in the robot and an output dimension of one. We will label the predicted cost-to-go at a state $\mathbf{x}$ as $\hat{J}(\mathbf{x}; \mathbf{\theta})$, where $\mathbf{\theta}$  are the trainable parameters of the net. 

\subsubsection{Loss Function}
\label{sec:hinge-loss}

We train the neural net from a set of training samples, where each sample consists of a tuple of (robot state $\mathbf{x}$, cost-to-go lower bound $J_{\text{lb}}$, and cost-to-go upper bound $J_{\text{ub}}$). We penalize the net for predicting values outside of the range $[J_\text{lb}, J_\text{ub}]$ using a double-sided hinge loss, defined as:
\begin{align}
h(\mathbf{x}, J_\text{lb}, J_\text{ub}; \mathbf{\theta}) = \begin{cases}
J_\text{lb} - \hat{J}(\mathbf{x}; \mathbf{\theta}) & \text{if } \hat{J}(\mathbf{x}; \mathbf{\theta}) < J_\text{lb}\\
0 & \text{if } J_\text{lb} \leq \hat{J}(\mathbf{x}; \mathbf{\theta}) \leq J_\text{ub} \\
\hat{J}(\mathbf{x}; \mathbf{\theta}) - J_\text{ub} & \text{if } \hat{J}(\mathbf{x}; \mathbf{\theta}) > J_\text{ub} \, .
\end{cases}
\label{eq:loss}
\end{align}
The total loss is simply the sum of the hinge losses over every sample $(\mathbf{x}, J_\text{lb}, J_\text{ub})$.

\begin{figure}[htpb]
  \centering
  \includegraphics[width=0.28\textwidth]{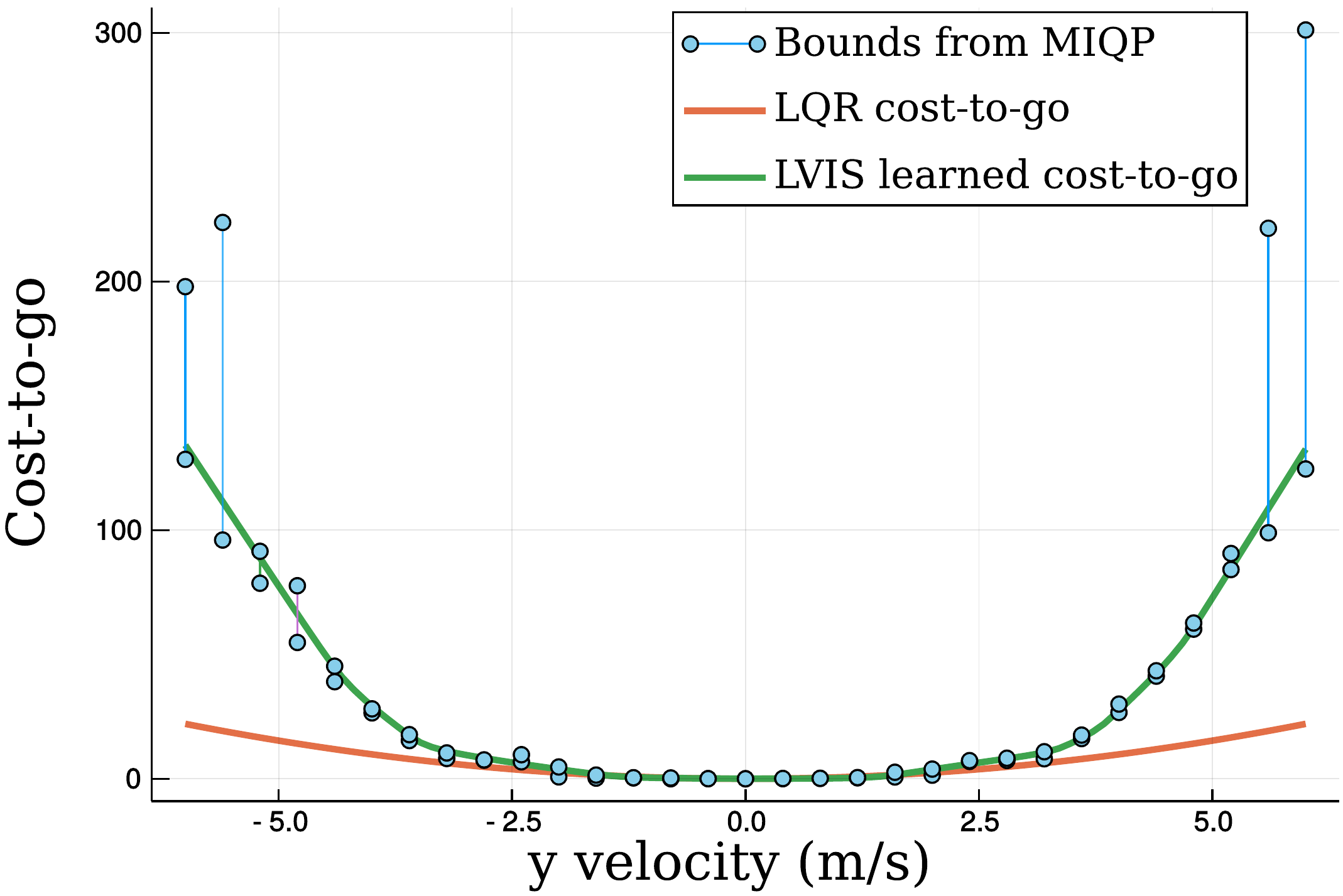}
  \caption{An example of an approximate value function learned from interval samples, as a function of the planar humanoid's initial $y$ velocity. Each connected pair of points represents the interval $[J_{lb}, J_{ub}]$ from a single trajectory optimization sample. For the states in which the mixed-integer program was solved to optimality, $J_{lb} \approx J_{ub}$. The prediction $\hat{J}(\mathbf{x}; \mathbf{\theta})$ in green lies within the interval $[J_{lb}, J_{ub}]$ at each sample. The sampled intervals match the LQR cost-to-go (red) for small initial velocities, indicating that the LQR policy is nearly optimal for small disturbances.}
  \label{fig:learned-cost-with-intervals}
\end{figure}

The use of the hinge loss provides a unique advantage: as training proceeds, we may revisit a particular state $\mathbf{x}$, and due to a better warm-start we may come up with tighter bounds $J_\text{lb}$ and $J_\text{ub}$ than we previously discovered. The hinge loss ensures that, so long as the net predicts a value $\hat{J}$ which falls within the new, tighter bounds, the old, looser bounds do not influence the total loss. If, instead, we were attempting to learn $J$ by exactly matching the upper or lower bounds or their precise midpoint, then our older samples would tend to pull the net's output away from the newer samples. Fig.~\ref{fig:learned-cost-with-intervals} shows an example of a learned value function and the interval samples which were used to train it.

\subsubsection{Optimization}

The parameters $\mathbf{\theta}$ are trained using a stock \textsc{Adam} optimizer, with all parameters set to the defaults suggested in \cite{KingmaAdamMethodStochastic2014}, and $\ell^2$ regularization was used to penalize the neural net weights. 

\subsection{Online Control Using the Learned Cost}
\label{sec:online}

The result of the training process described in Sect.~\ref{sec:training} is a neural net whose forward pass approximates the cost-to-go of the original MPC problem. To turn this neural net into a control policy, we construct a new MPC optimization with only one time step ($N=1$) and set as its terminal cost a local affine approximation of the neural net's predicted cost-to-go. This corresponds to a greedy gradient descent on the cost-to-go, subject to the robot's dynamics.

Even the one-step MPC optimization, however, still involves complementarity constraints and is thus a mixed-integer problem. The restriction to a single time step dramatically reduces the number of integer variables which must be solved, allowing near-real-time controller performance, but solving even these smaller MIQPs at control rates is still a challenge. For this work, we implemented a mixed-integer controller in Julia \cite{KoolenJuliaroboticssimulation2018} using the RigidBodyDynamics.jl software package to model the robot's dynamics \cite{rigidbodydynamicsjl}, the Parametron.jl software package to model the optimization problem \cite{KoolenParametronjl2018}, and the Gurobi optimizer to solve the resulting problems \cite{GurobiOptimizationInc.GurobiOptimizerReference2014}. 

\subsection{Choosing Initial States with \textsc{DAgger}}
\label{sec:dagger}

Rather than sampling randomly across the robot's entire state space, we adopt the \textsc{DAgger} approach from \cite{Rossreductionimitationlearning2011}. Put simply, \textsc{DAgger} relies on simulating the system using the (initially poorly-trained) policy as its controller instead of the expert, iteratively collecting new training samples from the regions of state-space visited by the learned policy. Our training alternates between (a) letting the learned controller drive the robot for 25-100 time steps while running the mixed-integer optimization to produce new training samples and (b) using those new samples to further train the approximate cost-to-go.

\subsection{Policy Net}
\label{sec:policy-net}

Rather than trying to learn the value function, we could simply attempt to train a neural net to mimic the mapping from $\mathbf{x}$ to $\mathbf{u}$ using the same trajectory optimization samples. We label this approach the Policy Net. As discussed in Sect.~\ref{sec:early-termination}, however, the fact that the trajectory optimizations are not generally solved to optimality means that the $\mathbf{u}$ samples are not generally optimal. Training a neural net to approximate these suboptimal samples is unlikely to result in a good approximation of the optimal policy, but we attempt to do so in order to evaluate that claim.
\section{RESULTS}



\subsection{Cart-Pole With Walls}
\label{sec:cart-pole-results}
The approximate cost-to-go for the cart pole was trained from 3862 mixed-integer trajectory optimization samples. Each trajectory optimization had a horizon of 20 steps and a time step of 25\,\si{ms}, for a total lookahead time of 0.5\,\si{\second}. Trajectory optimizations were terminated after 3 seconds, which was sufficient for 92.0\% of samples to converge to within 1\% of the globally optimal cost.
As a baseline policy, a discrete-time LQR controller was constructed using the linearization of the system about the upright configuration of the pole. The resulting LQR cost-to-go was used as the terminal cost during the offline mixed-integer trajectory optimization. 



\label{sec:cart-pole-training}

Training the cart-pole cost-to-go required approximately two hours on a single CPU, the majority of which was spent solving mixed-integer trajectory optimizations. A total of 500 rounds of training with the ADAM optimizer were performed. Convergence was estimated from 20\% of the training samples held as a validation set.

\begin{figure}[htb!]
  \centering
  \includegraphics[width=0.5\textwidth]{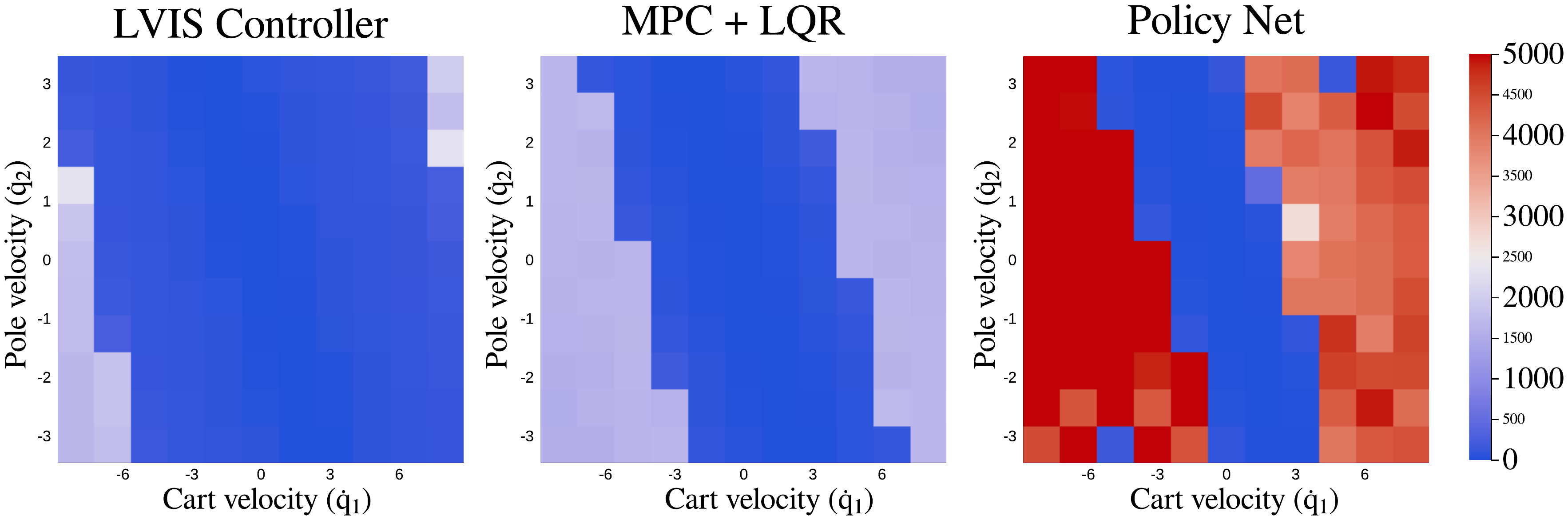}
  \caption{Accumulated cost of the cart-pole controllers. Each cell indicates the total accumulated cost over a 4-second simulation from the given initial cart and pole velocities, using the same cost matrices as the LQR controller. The regions of very low (dark blue) accumulated cost indicate simulations for which the pole was successfully balanced. The \lvis{} approach resulted in the lowest accumulated cost and successful stabilization from the widest variety of initial conditions.}
  \label{fig:cart-pole-simulation-grid}
\end{figure}


\subsubsection{Evaluation}

Three potential controllers were evaluated for the cart-pole in order to measure the effectiveness of the learned value function approach:
\begin{enumerate}
  \item \lvis{}: One-step mixed-integer MPC using the value function learned from the $[J_{lb}, J_{ub}]$ intervals as its terminal cost.
  \item MPC + LQR: One-step mixed-integer MPC using the LQR cost-to-go as its terminal cost. 
  \item Policy Net: The neural net trained to mimic the optimal policy (Sect.~\ref{sec:policy-net}).
\end{enumerate}

Each controller was evaluated by simulating the cart-pole for 4 seconds from a range of initial velocities. Each simulation began with the cart centered ($q_1 = 0$) and the pole upright ($q_2 = 0$), with initial cart velocity $\dot{q}_1$ ranging uniformly from -8\,\si[per-mode=symbol]{\metre\per\second} to 8\,\si[per-mode=symbol]{\metre\per\second} and initial pole rotational velocity $\dot{q}_2$ ranging uniformly from $-\pi\,\si[per-mode=symbol]{\radian\per\second}$ to $\pi\,\si[per-mode=symbol]{\radian\per\second}$. Eleven samples of each initial velocity were collected, for a total of 121 simulations of each controller. Performance of the controller was evaluated by measuring the total accumulated cost (using the same quadratic cost matrices $Q$ and $R$ that were used to design the LQR controller) over each simulation.


Results of the cart-pole simulation are shown in Fig.~\ref{fig:cart-pole-simulation-grid}. The \lvis{} controller out-performed the MPC + LQR baseline, resulting in a lower accumulated cost and successful stabilization of the pole from a wider range of initial velocities. The policy net controller (see Sect.~\ref{sec:policy-net}) was able to stabilize the pole from a few initial velocities, but it accumulated more cost than either of the MPC approaches in nearly every case. 


\subsection{Planar Humanoid}


For the planar humanoid, approximately 33,700 trajectory optimization samples were collected, with each trajectory optimization having a horizon of 10 and a time step of 50\,ms, for a total lookahead of 0.5\,s. Trajectory optimizations were terminated after 3\,\si{\second} of optimization with Gurobi. The higher state dimension and larger number of discrete modes made the humanoid trajectory optimization problems substantially harder to solve within that time limit, and only 3.2\% of optimizations could be solved to within 1\% of the optimal cost within that time.



\begin{figure}[htb!]
  \centering
  \includegraphics[width=0.48\textwidth]{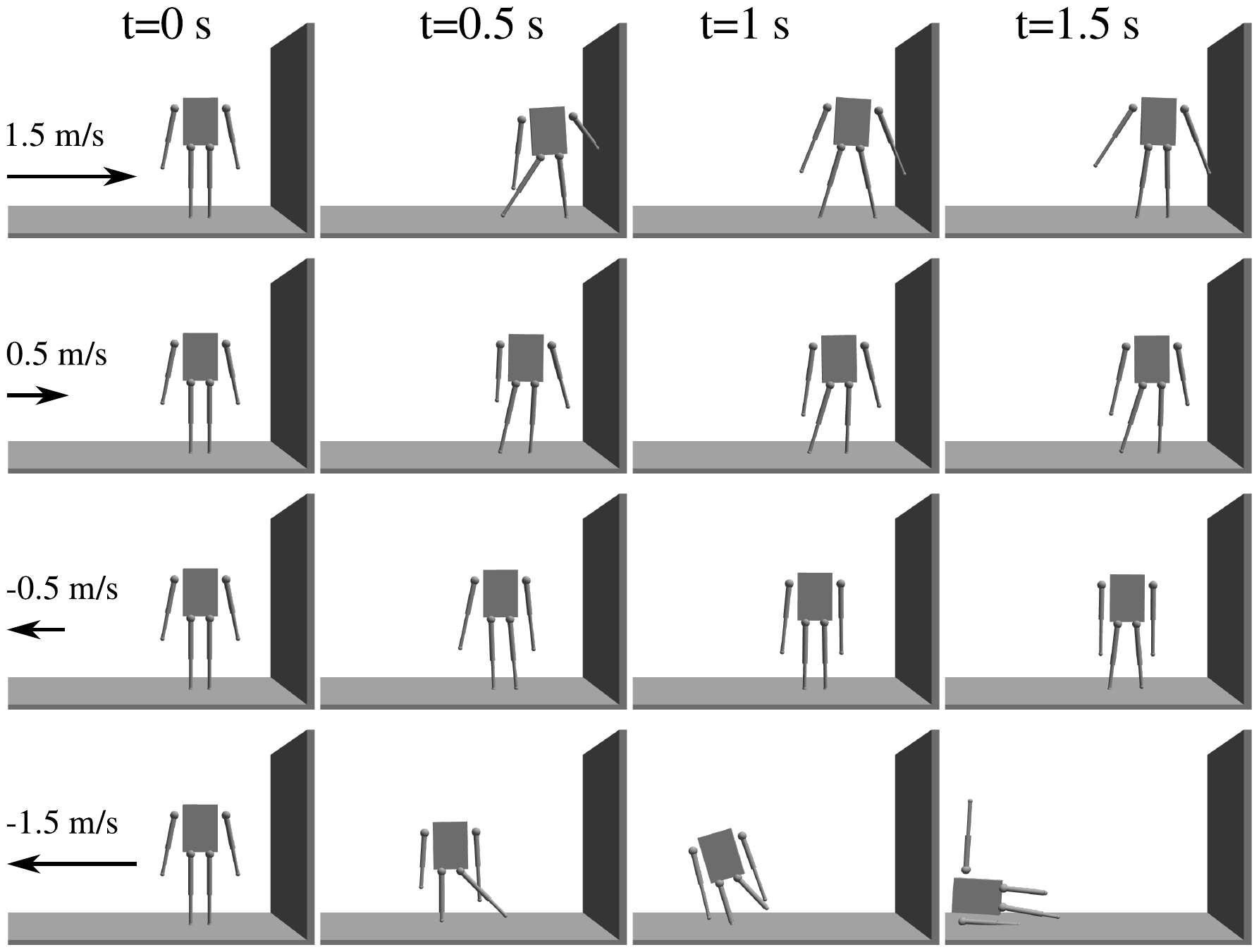}
  \caption{Animations of the planar humanoid recovering from pushes using the \lvis{} controller. Initial velocities refer to the velocity of the robot's torso along the $y$ axis of Fig.~\ref{fig:box-atlas}. Note that the robot can recover from a 1.5\,\si{\metre\per\second} velocity to the right by using contact with the wall, but cannot recover from the same initial velocity to the left.}
  \label{fig:box-atlas-animations}
\end{figure}

The method of Mason et al. \cite{MasonBalancingWalkingUsing2016} was used to generate an LQR policy consistent with the contact dynamics of the humanoid (the similar method of \cite{PosaOptimizationstabilizationtrajectories2016a} could also be used). The LQR policy was designed for the nominal configuration of the robot, shown in the left-most column of Fig.~\ref{fig:box-atlas-animations}, with both feet in contact with the ground.


Training the humanoid value function required approximately 36 CPU hours, again with the majority spent collecting trajectory optimization samples. A total of 300 rounds of training with the ADAM optimizer were performed, and convergence was estimated from 20\% of the training samples held as a validation set. 

\label{sec:policy-net-box-atlas}

A policy net was also trained on the humanoid optimization samples in an attempt to directly learn the mapping from state to action. The policy net also had two hidden layers with 48 units each, but had 11 outputs, corresponding to the 11 input dimensions of the robot. The same \textsc{DAgger} training process was run for the policy net, and the same 33,700 samples were provided for training.

\begin{figure}[htb!]
  \centering
  \includegraphics[width=0.5\textwidth]{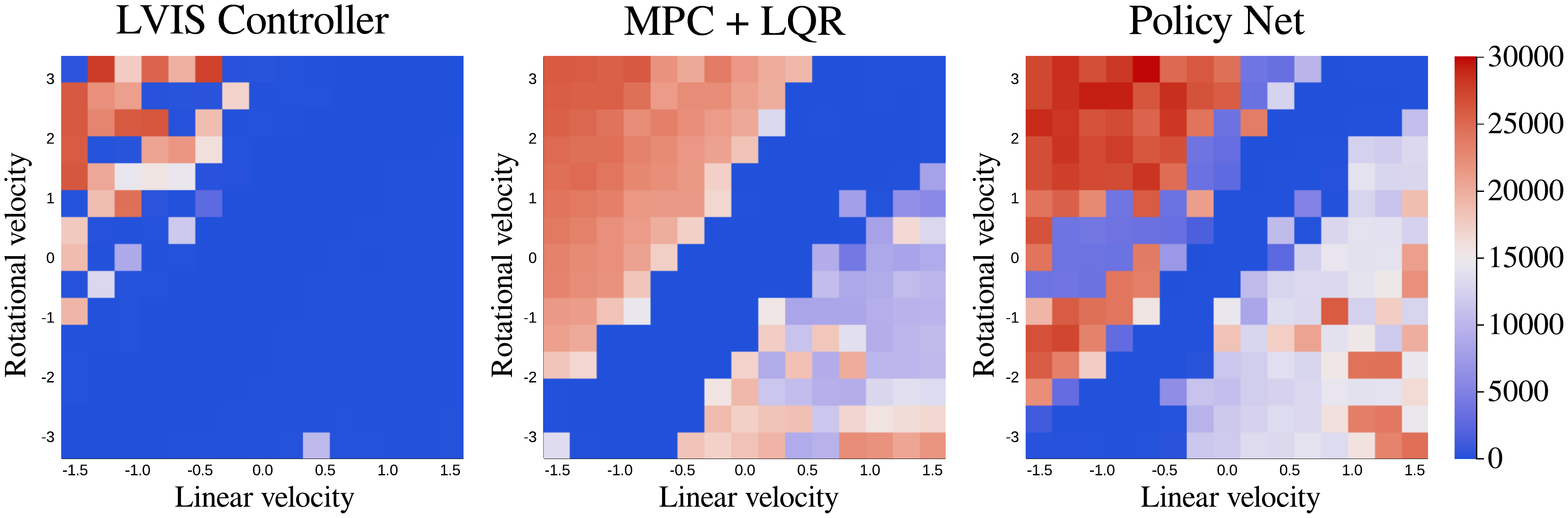}
  \caption{Accumulated cost of the humanoid controllers. Each cell indicates the total accumulated cost over a 4-second simulation from the given initial linear and angular velocity of the robot's body, using the same cost matrices as the LQR controller. \lvis{} achieved a low accumulated cost (dark blue) across a wide variety of initial conditions, performing particularly well in the bottom-right corner of the grid in which the initial velocity moved the robot towards the wall.}
  \label{fig:box-atlas-simulation-grid}
\end{figure}


\subsubsection{Evaluation}
\label{sec:box-atlas-evaluation}

The learned controller was evaluated by simulating the humanoid robot from a variety of initial velocities. From the nominal configuration, the robot's initial linear velocity (along the $y$ axis of Fig.~\ref{fig:box-atlas}) was varied from -1.5\,\si[per-mode=symbol]{\metre\per\second} to 1.5\,\si[per-mode=symbol]{\metre\per\second} and its initial angular velocity (about the $x$ axis of Fig.~\ref{fig:box-atlas}) was varied from $-\pi\,\si[per-mode=symbol]{\radian\per\second}$ to $\pi\,\si[per-mode=symbol]{\radian\per\second}$. The robot was then simulated under each control policy for 4 seconds using a simulated control rate of 100\,\si{Hz}. 


As was the case with the cart-pole, the \lvis{} controller generated substantially lower accumulated cost than the baseline controller using the LQR cost-to-go. In particular, the \lvis{} controller performed especially well when the robot's initial velocity directed it towards the wall, since it was able to both step and reach for the wall in order to maintain balance. The performance of the \lvis{} controller from a variety of initial velocities can be seen in Fig.~\ref{fig:box-atlas-animations}, and the learned cost is compared with LQR and the Policy Net in Fig.~\ref{fig:box-atlas-simulation-grid}. 

\begin{figure}[htb!]
  \centering
  \includegraphics[width=0.5\textwidth]{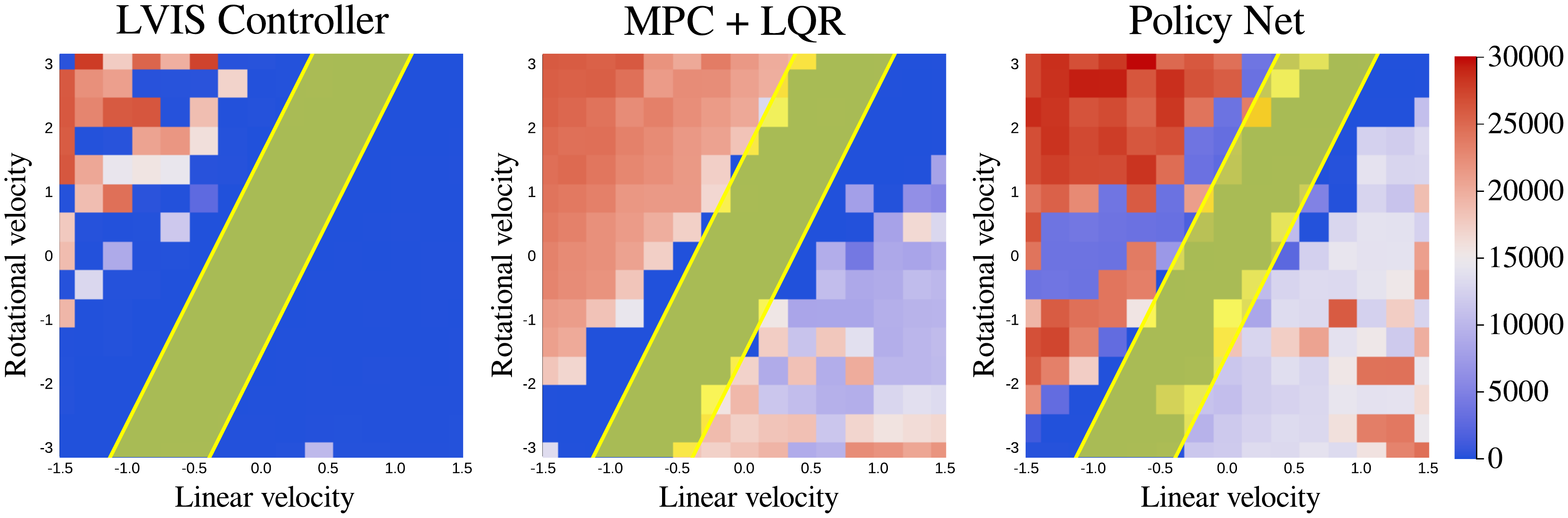}
  \caption{Comparing the performance of the humanoid controllers with the zero-step capturability region predicted by \cite{PrattCapturePointStep2006}. Each cell indicates the accumulated cost, as in Fig.~\ref{fig:box-atlas-simulation-grid}. For initial velocities in the yellow shaded region, the estimated Instantaneous Capture Point (ICP) lies between the robot's feet, so it should be possible for a controller to stabilize the center of mass without taking a step. The set of states stabilized by the LQR controller, indicated by the very low (dark blue) accumulated cost, approximately matches the region predicted by the ICP, while the \lvis{} controller stabilizes a much larger region.}
  \label{fig:box-atlas-capture-region}
\end{figure}


\subsubsection{Capturability Analysis}

Fig.~\ref{fig:box-atlas-simulation-grid} shows that the controller using the learned cost-to-go out-performs the baseline LQR controller, but it could simply be the case that the baseline LQR controller performed very poorly, making it easy to beat. To evaluate the performance of both controllers with respect to an independent benchmark, we can apply the Instantaneous Capture Point (ICP) work of Pratt et al. \cite{PrattCapturePointStep2006} to estimate the range of initial velocities for which the controller should be able to recover without taking a step. Fig.~\ref{fig:box-atlas-capture-region} shows in yellow the set of states for which the ICP predicts that the robot should be able to balance without taking a step. Even the baseline LQR controller is able to stabilize the robot from that entire range of states\footnote{While the LQR controller has no notion of stepping, it can sometimes successfully slide the foot in the direction of the initial velocity, resulting in stabilization even when the capture point predicts a fall.}, while the \lvis{} controller substantially out-performs the capture point predictions, since it is able to exploit changes in contact to stabilize the robot.

\begin{figure}[htb!]
  \centering
  \includegraphics[width=0.5\textwidth]{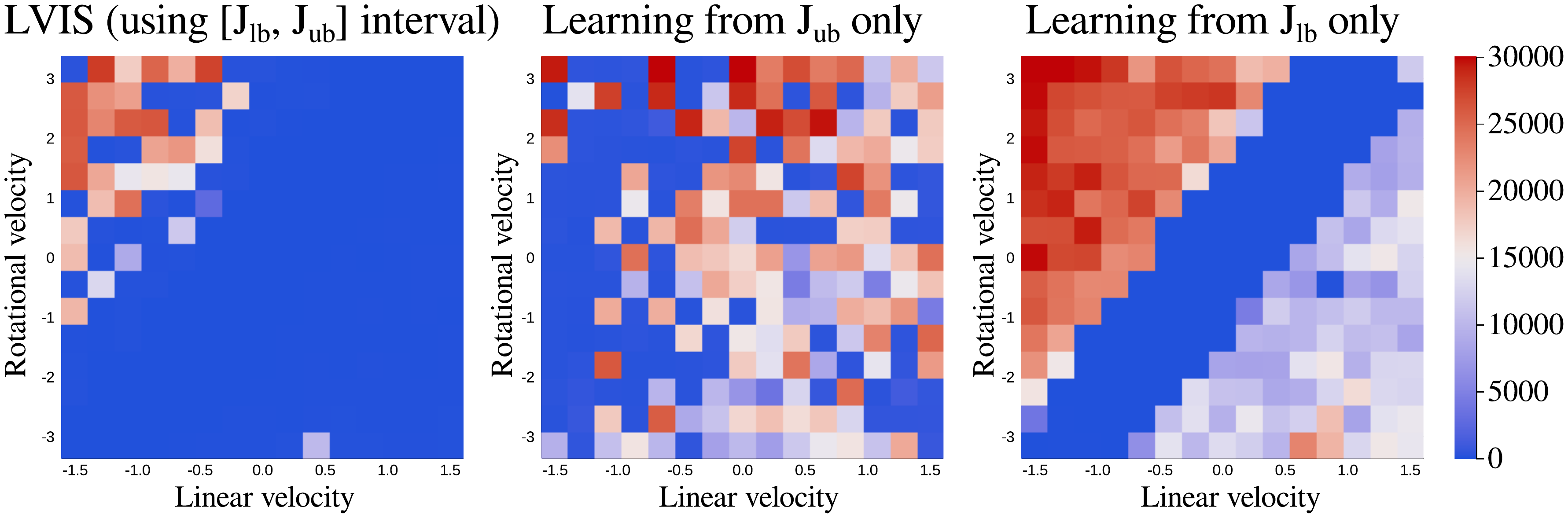}
  \caption{Comparing learning the cost-to-go from the bounded intervals (Sect.~\ref{sec:hinge-loss}), just the upper bound samples $J_{ub}$, or just the lower bound samples $J_{lb}$. Accumulated cost was measured as in Fig.~\ref{fig:box-atlas-simulation-grid}. Neither the upper nor the lower bounds alone were sufficient to train an approximate cost-to-go which out-performed the LQR baseline or the \lvis{} approach (left).}
  \label{fig:box-atlas-interval-vs-bounds}
\end{figure}


\subsubsection{The Importance of Intervals}
\label{sec:interval-importance}


As described in Sect.~\ref{sec:hinge-loss}, we only penalize the neural net for predicting a cost-to-go which is outside of the interval $[J_{lb}, J_{ub}]$. To test the validity of that approach, we also tried training the neural net to exactly mimic the best feasible value $J_{ub}$ or the best lower bound $J_{lb}$.
Two additional neural nets were trained using the same neural net structure, number of samples, and training process as in Sect.~\ref{sec:box-atlas-evaluation}. Each net was penalized for the $\ell^1$ error between its prediction and $J_{lb}$ or $J_{ub}$, respectively.
We evaluated both of these cost-to-go approximations using the same simulation procedure as in Fig.~\ref{fig:box-atlas-simulation-grid}. Neither the $J_{ub}$ samples alone nor the $J_{lb}$ samples alone produced a cost-to-go and controller which could out-perform even the LQR baseline, as shown in Fig.~\ref{fig:box-atlas-interval-vs-bounds}. In practice, attempting to train from just $J_{lb}$ or $J_{ub}$ resulted in substantial under-fitting, as the upper and lower bound data both showed a great deal of noise from one sample to the next, influenced by the quality of the warm-start solutions, and the varying behavior of Gurobi's internal heuristics.

\subsection{Learning in Parameterized Environments}
  
\begin{figure}[htb!]
  \includegraphics[width=0.38\textwidth]{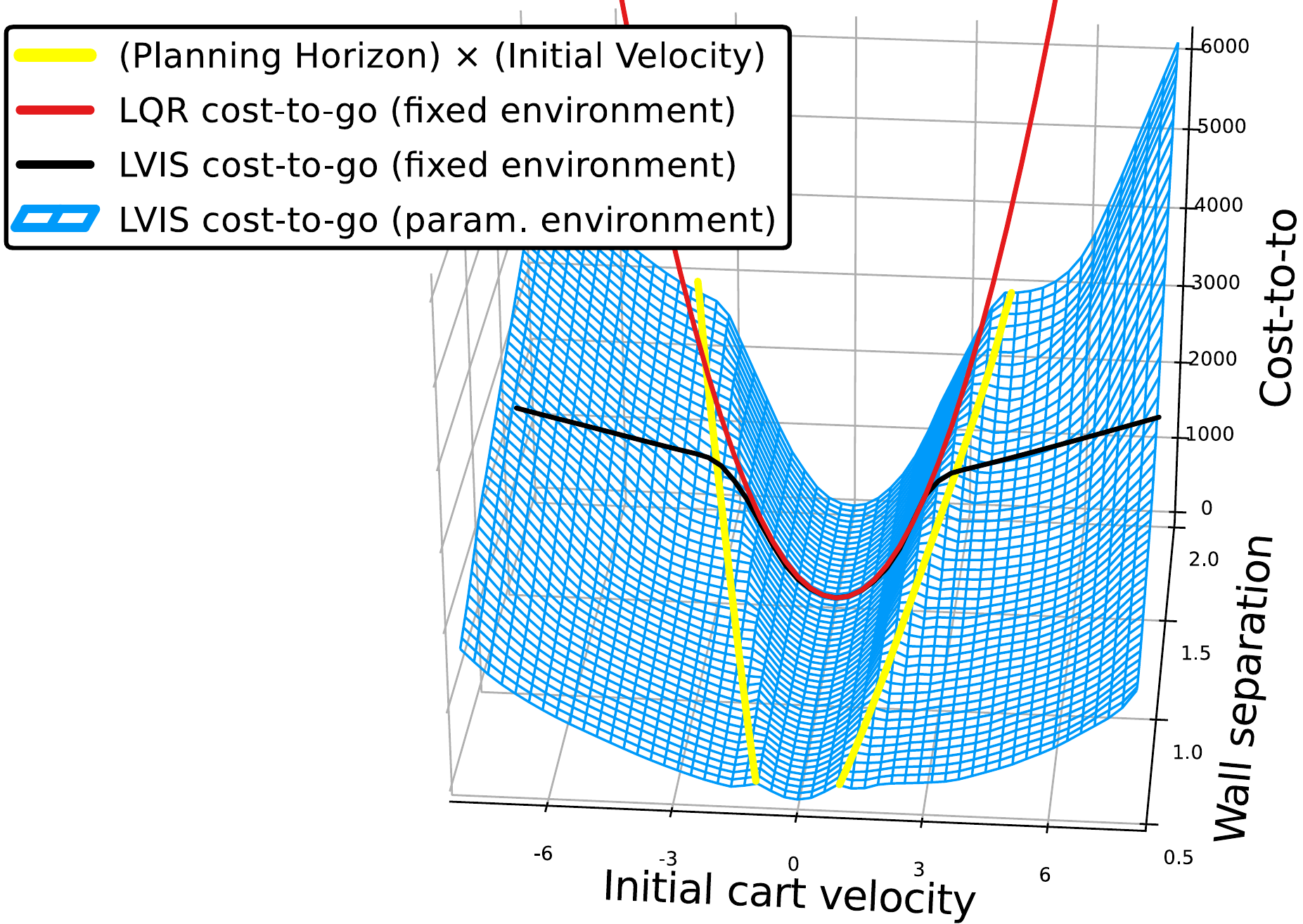}
  \caption{Learned cost-to-go (blue mesh) as a function of the initial cart velocity $\dot{q}_1$ and the parameter representing the distance from the center of the track to each wall. The parameterized cost-to-go closely matches LQR (in red) for states for which the cart will not reach the wall within the planning horizon (between the yellow lines), while it rises more gently elsewhere as the walls enable the robot to dissipate energy and reduce its cost. The black line shows the cost function learned in Sect.~\ref{sec:cart-pole-training}, for which the wall distance was fixed at 1.5\,\si{\metre}.}
  \label{fig:learned-param-cost-3d}
\end{figure}

One drawback of the \lvis{} approach is that the offline training and trajectory optimization are performed in a fixed environment. This essentially bakes that environment into the learned cost-to-go, resulting in a controller which is only useful in the trained environment. 
To handle a variety of environments, we suggest creating parameterized templates representing deformable environments. 
By encoding the environment parameters into the input to the \lvis{} neural net (both in training and at run-time), we can create a learned cost-to-go which is a function both of the robot's state and the environment parameters. 

As a basic demonstration of this approach, we modified the cart-pole environment (Fig.~\ref{fig:cart-pole}), adding a single parameter to represent the distance from the center of the track to the walls. The same training process as in Sect.~\ref{sec:cart-pole-training} was run, with 20,982 trajectory optimization samples collected over 18 hours. For each iteration of the \textsc{DAgger} training, the distance to the walls was uniformly randomly varied from 0.5\,\si{\metre} to 2.0\,\si{\metre}. The resulting learned cost-to-go (now a function of $\mathbf{x}$ and that parameter) is shown in Fig.~\ref{fig:learned-param-cost-3d}. 

\section{FUTURE WORK}
\label{sec:future-minlp}
The most significant limitation of this work is the use of the linearized dynamics, which means that the learned cost-to-go is only valid for that linearization. We believe, however, that this limitation can be overcome: there exist spatial branch-and-bound techniques, analogous to the tools used to solve MIQPs, which are applicable to general nonlinear mixed-integer optimizations. These techniques can also provide rigorous upper and lower bounds on the cost-to-go with no need to linearize. We look forward to experimenting with the full nonlinear dynamics, using tools like Couenne \cite{BelottiBranchingboundstighteningtechniques2009} or BARON \cite{Tawarmalanipolyhedralbranchandcutapproach2005} to solve the resulting optimizations. 

\addtolength{\textheight}{-10cm}   

\bibliographystyle{latex/IEEEtran}
\bibliography{refs}

\end{document}